\documentclass[cameraready]{Interspeech}
\title{SCRIBE: Diagnostic Evaluation and Rich Transcription Models for Indic ASR}
\author[orcid=0000-0003-2402-5272]{Kavya}{Manohar}
\author[orcid=0009-0007-3192-4563]{Arghya}{Bhattacharya}
\author[orcid=0009-0000-8240-0099]{Kush}{Juvekar}
\author[orcid=0009-0003-0730-6897]{Kumarmanas}{Nethil}

\address{
    Adalat AI, India
}
\email{\{kavya,arghya,kush,manas\}@adalat.ai}

\usepackage{comment}
\usepackage{hyperref}
\usepackage{url}
\usepackage{tikz}
\usetikzlibrary{shapes.geometric, arrows, positioning, calc, backgrounds, fit, shadows}
\usepackage{fontspec}
\usepackage{polyglossia}
\usetikzlibrary{positioning, shapes.misc, fit, backgrounds}
\setmainlanguage{english}
\setotherlanguages{hindi, malayalam}
\newfontfamily\hindifont{NotoSansDevanagari-Regular.ttf}[Script=Devanagari]
\newfontfamily\malayalamfont{NotoSansMalayalam-Regular.ttf}[Script=Malayalam, Scale=0.9]
\newfontfamily\ipafont{CharisSIL-Regular.ttf}[Scale=MatchLowercase]  
\usepackage{newtxtext}
\newcommand{\ipa}[1]{{\ipafont\scriptsize #1}}
\usepackage[table]{xcolor} 
\usepackage{array}         
\usepackage{multirow}
\colorlet{lightgreen}{green!20}

\newcolumntype{g}{>{\columncolor{lightgray}}c}

\usepackage{booktabs} 
\usepackage{multirow} 
\usepackage{amsmath}  
\begin{document}
\maketitle

\begin{abstract}

Automatic speech recognition replaces typing only when correction costs less than manual entry, a threshold determined by error types, not counts: fixing a misrecognized domain term costs far more than inserting a comma. Word error rate (WER) fails on two fronts: it collapses distinct error categories into a single scalar, and it structurally penalizes agglutinative languages where valid \textit{sandhi} merges inflate scores. We introduce SCRIBE, a diagnostic framework that provides categorical error decomposition into lexical, punctuation, numeral, and domain-entity rates through \textit{sandhi}-tolerant alignment with domain vocabulary injection. Human validation confirms SCRIBE aligns with expert judgment where WER does not. We release SCRIBE, an LLM curation pipeline, benchmarks, and open-weight rich transcription models for Hindi, Malayalam, and Kannada.

\end{abstract}

\section{Introduction}

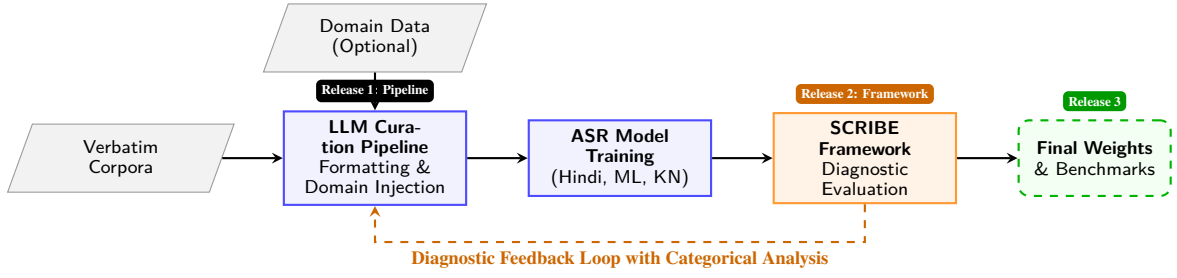
\begin{figure*}[t]
\centering
\begin{tikzpicture}[
    >=stealth,
    node distance=0.6cm and 0.8cm,
    font=\sffamily\scriptsize,
    data/.style={trapezium, trapezium left angle=70, trapezium right angle=110, draw=gray!80, fill=gray!10, text width=1.6cm, align=center, minimum height=0.9cm},
    process/.style={rectangle, draw=blue!70, fill=blue!5, text width=2.2cm, align=center, minimum height=1.0cm, thick},
    scribe/.style={rectangle, draw=orange!80, fill=orange!10, text width=2.2cm, align=center, minimum height=1.0cm, thick},
    release/.style={rectangle, rounded corners, draw=green!70!black, fill=green!5, text width=1.8cm, align=center, minimum height=1.0cm, thick, dashed},
    tag/.style={rectangle, fill=black, text=white, font=\bfseries\tiny, inner sep=2pt, rounded corners=2pt}
]
    \node[data] (input) {Verbatim \\ Corpora};
    \node[process, right=of input] (curation) {\textbf{LLM Curation Pipeline} \\ \scriptsize Formatting \& Domain Injection};
    \node[tag, above=0.1cm of curation] {Release 1: Pipeline};
    \node[process, right=of curation] (training) {\textbf{ASR Model Training} \\ \scriptsize (Hindi, ML, KN)};
    \node[scribe, right=of training] (eval) {\textbf{SCRIBE Framework} \\ \scriptsize Diagnostic Evaluation};
    \node[tag, above=0.1cm of eval, fill=orange!80!black] {Release 2: Framework};
    \node[release, right=of eval] (models) {\textbf{Final Weights} \\ \scriptsize \& Benchmarks};
    \node[tag, above=0.1cm of models, fill=green!60!black] {Release 3};
    \node[data, above=0.5cm of curation] (domain) {Domain Data (Optional)};
    \draw[->, thick] (input) -- (curation);
    \draw[->, thick] (curation) -- (training);
    \draw[->, thick] (training) -- (eval);
    \draw[->, thick] (eval) -- (models);
    \draw[->, thick] (domain) -- (curation);
    \draw[->, thick, dashed, draw=orange!80!black] (eval.south) -- ++(0,-0.5) -|
        node[pos=0.25, below, font=\bfseries\scriptsize, text=orange!80!black] {Diagnostic Feedback Loop with Categorical Analysis}
        (curation.south);
\end{tikzpicture}
\caption{Diagnostic-led development cycle for Indic rich transcription. SCRIBE provides the categorical feedback necessary to refine curation and verify model performance across error types.}
\label{fig:pipeline}
\vspace{-0.4cm}
\end{figure*}

The utility of automatic speech recognition (ASR) for dictation, producing medical notes, legal proceedings, or classroom transcripts, is defined by the correction threshold: editing must be faster than typing. This requires \textit{rich transcription}: text with grammatical punctuation, standardized numerals, and domain-appropriate orthographic conventions. Whether a system meets this bar depends on the \textit{type} of error, not just the count. A missing comma is trivial; a misrecognized medical term or incorrectly formatted legal date can render output unusable.

Standard word error rate (WER) fails as a development signal for two reasons. First, it collapses acoustic failures, numeral formatting, and punctuation into a single scalar, offering no actionable insight. Second, it is structurally broken for agglutinative Indic languages. In morphologically complex Dravidian languages like Malayalam and Kannada \cite{manohar2020quantitative, bharadwaja2007statistical}, valid word-boundary merges (\textit{sandhi}) with phonotactic changes at the boundaries trigger cascading alignment shifts in 1:1 alignment, inflating error rates by up to 30\% relative. This is a structural penalty against an entire language family.

We introduce SCRIBE, a diagnostic evaluation framework named for the role it measures: whether ASR can serve as a reliable scribe. Rather than a single scalar, SCRIBE outputs a diagnostic error vector $\mathbf{E} = [ER_{lex}, ER_{punc}, ER_{num}, ER_{ent}]$, which decomposes failures into lexical, punctuation, numeral, and domain-specific entity error rates, respectively. By utilizing sandhi-tolerant alignment and categorical decomposition, SCRIBE augments the monolithic WER with an actionable development signal, enabling targeted remediation for rich transcription tasks.

\noindent To summarize, our major contributions in this paper are:
\begin{itemize}
    \item \textbf{SCRIBE}\footnote{\url{https://github.com/adalat-ai-tech/scribe-eval}}, released as an open-source evaluation tool, providing \textit{sandhi}-tolerant alignment and categorical error decomposition, proposed as a diagnostic complement to monolithic WER wherever ASR serves as a scribe.
    \item \textbf{A structured annotation schema and validation procedure} for categorical ASR metrics, with dimension-specific scales rated independently by expert linguists, demonstrating that SCRIBE aligns with human judgment where WER does not.
    \item \textbf{A reproducible recipe for Indic rich transcription}\footnote{\url{https://huggingface.co/collections/adalat-ai/scribe-dictation-models-and-benchmarks}}: an LLM-based data curation pipeline, two new benchmarks (FLEURS-RO for general and IN22-Legal for domain evaluation), and the first open-weight rich transcription models for Hindi, Malayalam, and Kannada.
\end{itemize}

\section{Related Work}

\textbf{Rich Transcription Models:} While models like Whisper \cite{radford2023robust} and Canary \cite{raokoluguri25_interspeech} demonstrate the feasibility of joint acoustic-orthographic modeling, the open-source Indic ecosystem remains dominated by verbatim-only models \cite{bhogale2023effectiveness, bhogale23_interspeech, bhogale2025towards}. Current pipelines for formatted output often rely on decoupled inverse text normalization \cite{pulipaka2025mark}, which ignores prosodic cues and homophone resolution. SCRIBE bridges this by providing a recipe for native rich transcription in Indic ASR.

\textbf{Evaluation Limitations:} While character error rate (CER) sidesteps word-boundary issues in agglutinative languages \cite{k-etal-2025-advocating}, it and WER lack diagnostic signals. CER is semantically blind, weighting functional suffix changes identically to root morpheme substitutions. Categorical frameworks like Beyond Levenshtein \cite{kuhn24_interspeech} move toward nuanced evaluation but rely on normalization that destructively strips lexically indispensable Indic vowel signs (\textit{matras}) and diacritics \cite{manohar-pillai-2024-lost}. Similarly, 1:1 word alignment—shared by word information lost (WIL) and word information preserved (WIP) \cite{morris04_interspeech}—cannot resolve valid \textit{sandhi} (word-boundary merges) common in Dravidian languages \cite{manohar-pillai-2024-lost}. Semantic metrics like Semascore \cite{semascore2024} prioritize global meaning but remain blind to fatal numeral or negation failures that render professional dictation unusable. While the Orthographically-Informed WER leverages LLMs to capture permissible variations \cite{bhogale2026orthographicallyinformedevaluationspeechrecognition}, the approach is computationally expensive and fails to resolve the structural alignment shifts caused by \textit{sandhi}. SCRIBE addresses these gaps through diacritic-preserving normalization, deterministic \textit{sandhi}-tolerant alignment, and categorical error decomposition.

\section{The SCRIBE Framework}
\label{sec:framework}

\textsc{SCRIBE} is organized into three phases: tokenization and domain shielding, a sandhi-aware alignment engine, and categorical error aggregation. The framework outputs a diagnostic error vector $\mathbf{E}$ where each component maps to a specific remediation strategy.

\subsection{Phase 1: Tokenization and Domain Shielding}

 The framework transforms reference $R$ and hypothesis $H$ into typed tokens $(w_i, t_i)$, where $t_i \in \{\text{lexeme, numeral, punctuation, domain-entity}\}$. Unlike standard tokenizers that strip or blindly isolate punctuation, in SCRIBE standard punctuation and Indic-specific marks (e.g., the Hindi \textit{danda}) become independent tokens, while punctuation within numerals and compound words (e.g., \texttt{22.05.2023}, \texttt{ice-cream}) are preserved to maintain lexical integrity. User-defined domain entities are injected via a regex-based shielding layer to treat them as atomic units, preventing spurious fragmentation.



\subsection{Phase 2: Sandhi-Aware Alignment Engine}

An \emph{alignment} is a pairing of reference, $R$ and hypothesis, $H$ positions that accounts for insertions, deletions, standard 1:1 substitutions, and \textit{Sandhi}-motivated 1:2 (split) and 2:1 (merge) mappings. We seek the alignment maximizing a total score $dp[i][j]$, calculated via extended dynamic programming in Equation \ref{eq:recurrence}.

\begin{equation}
\label{eq:recurrence}
dp[i][j] = \max \left\{
\begin{aligned}
& dp[i\text{-}1][j\text{-}1] + S(r_i, h_j) && \text{(match/sub)} \\
& dp[i\text{-}1][j] + \gamma(t^R_i)   && \text{(deletion)} \\
& dp[i][j\text{-}1] + \gamma(t^H_j)   && \text{(insertion)} \\
& dp[i\text{-}1][j\text{-}2] + \Sigma_{\text{split}} && \text{(Sandhi-split)} \\
& dp[i\text{-}2][j\text{-}1] + \Sigma_{\text{merge}} && \text{(Sandhi-merge)}
\end{aligned}
\right.
\end{equation}



 

The scoring function $S(r_i, h_j)$ anchors the alignment on exact matches ($\alpha = +4.0$) while buffering against the acoustic near-misses common in Indic scripts (e.g., \texthindi{खाना}:\textit{khana} vs \texthindi{गाना}:\textit{gana}). To prevent alignment drift, a category-clash penalty $\beta = -3.0$ is applied if $t^R_i \neq t^H_j$. For same-category substitutions, we employ a Levenshtein-buffered penalty $\delta = -1.5 - (0.2 \cdot d)$, where $d$ is the character distance between $r_i$ and $h_j$. These coefficients were fixed via a test-driven procedure—tuned against expert-curated alignment cases until the engine reproduced the intended ordering (cross-category penalties exceeding same-category ones)—and are exposed as configurable parameters in the released library for re-tuning on new languages. Sensitivity analysis on our target languages confirms that a Unicode-level distance of $d \leq 2$ optimally captures minor orthographic variations—such as \textit{matra} shifts or gemination—without triggering the cascading deletion-insertion pairs typical of standard WER evaluation.

Sandhi scores, $\Sigma$, resolve 1:2 or 2:1 mappings by validating phonetic plausibility. A transition is valid if the fused form $s$ matches the prefix of $w_1$ and suffix of $w_2$. We score this as $\Sigma = \alpha + \sigma - d(b_{split}, b_{mid})/|s|$, where $\sigma = -0.5$ is the sandhi penalty. If the boundary distance $d > 2$, the transition is invalidated per Indic morphophonological rules \cite{bhardwaj-etal-2018-sandhikosh, dasari-etal-2025-sandhi, thottingal-2019-finite}. Figure \ref{fig:alignment_comparison} illustrates SCRIBE's ability to correctly resolve these complex word merges.

\begin{figure}[h]
\vspace{-0.2cm}

\centering
\begin{tikzpicture}[x=1.6cm, y=-0.9cm, font=\footnotesize]
  \node[anchor=west] at (-0.4, -0.8) {\textbf{SCRIBE Sandhi-Aware Alignment}};
  \node (dr1) at (0,0) {\textmalayalam{ഇന്ന്}};
  \node[anchor=south, inner sep=0pt] at ([yshift=0pt]dr1.north) {\ipa{[in̪n̪ə]}};
  \node (dr2) at (0.8,0) {\textmalayalam{അല്ലെങ്കിൽ}};
  \node[anchor=south, inner sep=0pt] at ([yshift=0pt]dr2.north) {\ipa{[alleŋkil]}};
  \node (dr3) at (2.2,0) {\textmalayalam{നാളെയാകട്ടെ}};
  \node[anchor=south, inner sep=0pt] at ([yshift=0pt]dr3.north) {\ipa{[n̪aːɭejaːkaʈʈe]}};
  \node (dh1) at (0.4,1.4) {\textmalayalam{ഇന്നല്ലെങ്കിൽ}};
  \node[anchor=north, inner sep=0pt] at ([yshift=0pt]dh1.south) {\ipa{[in̪n̪alleŋkil]}};
  \node (dh2) at (1.8,1.4) {\textmalayalam{നാളെ}};
  \node[anchor=north, inner sep=0pt] at ([yshift=0pt]dh2.south) {\ipa{[n̪aːɭe]}};
  \node (dh3) at (2.6,1.4) {\textmalayalam{ആകട്ടെ}};
  \node[anchor=north, inner sep=0pt] at ([yshift=0pt]dh3.south) {\ipa{[aːkaʈʈe]}};
  \draw[thick, blue!80!black] (dr1.south) -- (dh1.north);
  \draw[thick, blue!80!black] (dr2.south) -- (dh1.north);
  \node[blue!80!black, fill=white, inner sep=1pt, font=\bfseries\tiny] at (0.4, 0.7) {MERGE (2:1)};
  \draw[thick, orange!80!black] (dr3.south) -- (dh2.north);
  \draw[thick, orange!80!black] (dr3.south) -- (dh3.north);
  \node[orange!80!black, fill=white, inner sep=1pt, font=\bfseries\tiny] at (2.2, 0.7) {SPLIT (1:2)};
  \node[anchor=north, font=\scriptsize\itshape, color=blue!80!black] at (1.3, 2.2) {$E_{lex}=0\%$: Correct resolution of linguistic merges and splits};
  \draw[thick, gray!20] (-0.6, -1.5) -- (3.2, -1.5);
  \begin{scope}[yshift=3.8cm]
  \node[anchor=west] at (-0.6, -0.8) {\textbf{Standard 1:1 Alignment (JIWER)}};
  \node (sr1) at (0,0) {\textmalayalam{ഇന്ന്}};
  \node[anchor=south, inner sep=0pt] at ([yshift=0pt]sr1.north) {\ipa{[in̪n̪ə]}};
  \node (sr2) at (1,0) {\textmalayalam{അല്ലെങ്കിൽ}};
  \node[anchor=south, inner sep=0pt] at ([yshift=0pt]sr2.north) {\ipa{[alleŋkil]}};
  \node (sr3) at (2.2,0) {\textmalayalam{നാളെയാകട്ടെ}};
  \node[anchor=south, inner sep=0pt] at ([yshift=0pt]sr3.north) {\ipa{[n̪aːɭejaːkaʈʈe]}};
  \node (sh1) at (0,1.4) {\textmalayalam{ഇന്നല്ലെങ്കിൽ}};
  \node[anchor=north, inner sep=0pt] at ([yshift=0pt]sh1.south) {\ipa{[in̪n̪alleŋkil]}};
  \node (sh2) at (1,1.4) {\textmalayalam{നാളെ}};
  \node[anchor=north, inner sep=0pt] at ([yshift=0pt]sh2.south) {\ipa{[n̪aːɭe]}};
  \node (sh3) at (2.2,1.4) {\textmalayalam{ആകട്ടെ}};
  \node[anchor=north, inner sep=0pt] at ([yshift=0pt]sh3.south) {\ipa{[aːkaʈʈe]}};
  \draw[thick, red] (sr1) -- (sh1) node[midway, fill=white, inner sep=1pt, scale=0.8] {sub};
  \draw[thick, red] (sr2) -- (sh2) node[midway, fill=white, inner sep=1pt, scale=0.8] {sub};
  \draw[thick, red] (sr3) -- (sh3) node[midway, fill=white, inner sep=1pt, scale=0.8] {sub};
  \node[anchor=north, font=\scriptsize\itshape, color=red] at (1.1, 2.2) {$WER=100\% $: Alignment shift due to word-splits and merges};
  \end{scope}
\end{tikzpicture}
\caption{Standard libraries trigger cascading alignment shifts during linguistic merges and splits, inflating the WER, whereas SCRIBE correctly identifies these orthographic variations reporting 0\% $ER_{lex}$.}
\label{fig:alignment_comparison}
\vspace{-0.4cm}
\end{figure}

\subsection{Phase 3: Categorical Error Aggregation}
\label{sec:phase3}

SCRIBE aggregates errors into a diagnostic vector $\mathbf{E} = [ER_{lex}, ER_{punc}, ER_{num}, ER_{ent}]$. Each rate uses a combined denominator $N_{\text{comb}} = \sum_{t \in \mathcal{T}} \text{total}[t]$: $ER_t = (sub[t] + ins[t] + del[t]) / N_{\text{comb}}$. This shared scaling keeps sparse-category failures from producing misleadingly high rates and makes the components directly summable: their sum, which we term \emph{SCRIBE-WER} ($\mathrm{WER_S} = \sum_t ER_t$), is a single global rate that serves the summarising role of WER with sandhi inflation removed. Table~\ref{tab:combined_results} reports standard WER alongside the components from which $\mathrm{WER_S}$ is recovered, so SCRIBE extends WER rather than replacing it---preserving the global view while exposing the per-category decomposition beneath it. To respect valid formatting choices in professional dictation, it optionally normalizes date and numeral delimiters so that acceptable orthographic variations do not inflate $ER_{num}$. It also emits per-category reports with absolute error counts for targeted remediation.

\section{Experimental Setup}
\label{sec:setup}

We validate SCRIBE through a complete experimental cycle of rich transcription model development for Hindi, Malayalam, and Kannada (Figure~\ref{fig:pipeline}). This section describes: (1)~the LLM-based data curation pipeline and the rich transcription models trained on it; (2)~two new benchmarks released for general and domain-specific evaluation; and (3)~a human evaluation study with expert linguists designed to test whether SCRIBE's categorical rates align with human judgment where monolithic WER does not.

\subsection{Data and Models}

Public Indic speech corpora \cite{bhogale2023effectiveness,kathbath2022, prahallad2012iiit, gopinath2022imasc, javed2024indicvoices, baby2016resources, conneau2023fleurs} provide mostly verbatim transcripts. We use Gemini 2.5 Pro~\cite{comanici2025gemini} with language-specific prompts to transform these into rich transcription. A multi-tier quality control pipeline discards samples where CER exceeds thresholds for lexical changes (ignoring numeral and punctuation shifts) or where foreign-script characters are detected, removing $\sim$10\% of data.

The final curated sets comprise $\sim$1000h Hindi, $\sim$850h Kannada, and $\sim$800h Malayalam. SCRIBE-ASR is fine-tuned from a pre-trained Whisper-small and Whisper-medium architecture in three stages: (1)~diversity adaptation across acoustic conditions, (2)~pace and style robustness, and (3)~precision tuning with near perfect well articulated speech. We compare against two baselines: IndicWhisper (Vistaar) \cite{bhogale2023effectiveness} and IndicConformer \cite{javed2024indicvoices}, neither of which claims rich transcription natively.

\subsection{Benchmarks}

Existing Indic ASR benchmarks evaluate only verbatim transcription and offer no way to measure formatting accuracy. We release two curated evaluation sets designed to fill this gap across general and domain-specific conditions.

\textbf{FLEURS-RO} (Rich Orthography) is derived from the FLEURS multilingual test set \cite{conneau2023fleurs}. We apply our LLM curation pipeline to generate rich transcription references for the Hindi, Kannada, and Malayalam splits. Each transformed reference is then verified by a native-speaker linguist who corrects hallucinated punctuation, numeral formatting errors, and script inconsistencies introduced by the LLM. The result is a general-domain benchmark where both verbatim and rich transcription ground truths are available.

\textbf{IN22-Legal} is a domain-specific out-of-distribution benchmark derived from IN22 \cite{gala2023indictrans2}. Legal passages were recorded as read speech by 2--4 speakers per language ($\sim$30 minutes per language), producing a corpus dense in domain entities (statute names, section numbers), formal numerals (dates, monetary amounts), and complex clause structures. Ground-truth transcripts were prepared directly in rich transcription format by legal-domain annotators. Because none of the training data contains legal text, IN22-Legal tests whether formatting conventions learned from general corpora generalize to high-stakes specialized vocabulary.

\subsection{Human Evaluation Protocol}
\label{sec:human_eval}

The central claim of SCRIBE is that categorical error rates capture distinctions that experts perceive but monolithic WER cannot. To test this, we design a correlation study where human ratings serve as the ground truth against which both SCRIBE and WER are measured. If SCRIBE's per-category rates correlate significantly more strongly with expert judgment than WER does, the decomposition is validated as a meaningful diagnostic signal.

\textbf{Annotators and samples.} We selected 80 samples per language (240 total) from the IN22-Legal benchmark to ensure high density of domain-specific entities, numerals, and complex punctuation. Eight expert linguists (two per language), each a native speaker with professional proficiency in formal written registers, independently rated the SCRIBE-ASR hypotheses against ground-truth transcripts.

\textbf{Annotation schema.} Annotators assign scores on a 1.0--5.0 continuous scale (decimal scores encouraged for fine-grained discrimination) across three dimensions:
\begin{itemize}
    \item \textbf{Lexical accuracy (S1):} Correctness of base words, evaluated independently of formatting. 5.0 = every spoken word present and correct; 3.0 = meaning preserved with 2--3 errors; 1.0 = wholesale misrecognition.
    \item \textbf{Numeral accuracy (S2):} Correctness and format compliance of numbers and dates. 5.0 = mathematically accurate with proper digit formatting (e.g., ``302'' not ``three hundred two''); 1.0 = mathematically incorrect values (e.g., Section 302 $\to$ Section 307), constituting fatal errors in legal contexts.
    \item \textbf{Punctuation accuracy (S3):} Appropriateness of sentence boundaries, commas, and Indic-specific marks (e.g., \textit{danda}). 5.0 = professional-grade segmentation; 1.0 = absent or misleading punctuation.
\end{itemize}

We use a continuous rather than discrete scale because Spearman correlation requires sufficient rank variation; a coarse 3-point scale would compress distinctions that experts naturally perceive (e.g., one misplaced comma vs.\ five). Dimensions are rated independently to prevent halo effects: annotators complete all S1 ratings before proceeding to S2, ensuring that a strong lexical impression does not inflate punctuation scores. Samples where a category is absent (e.g., no numerals) are marked \texttt{N/A} and excluded from that category's correlation. Annotators were calibrated via written guidelines with worked examples distinguishing minor formatting variances (e.g., comma placement preference) from fatal value errors (e.g., wrong statute number), and recognizing valid \textit{sandhi} variations that should not be penalized.

\textbf{Analysis.} We compute Spearman $\rho$ between SCRIBE's categorical error rates ($ER_{lex}$, $ER_{num}$, $ER_{punc}$) and their corresponding human dimensions ($S1$, $S2$, $S3$), and contrast these against the correlation of monolithic WER with each dimension. 

\begin{table*}[t]
\centering
\caption{\textsc{SCRIBE} decomposition on general and legal benchmarks. All values are error rates (\%); $\mathrm{WER_S}=\sum_t ER_t$ is the aggregate SCRIBE rate (Sec.~\ref{sec:phase3}). WER is the value obtained from the JIWER library. Best per language in \textbf{bold}.}
\label{tab:combined_results}
\setlength{\tabcolsep}{3.5pt}
\begin{tabular}{ll gggg c ggggg c}
\toprule
& & \multicolumn{5}{c}{\textbf{FLEURS-RO (General)}} & \multicolumn{6}{c}{\textbf{IN22-Legal (Domain Specific)}} \\
\cmidrule(lr){3-7} \cmidrule(lr){8-13}
\textbf{Lang.} & \textbf{Model} & $ER_{lex}$ & $ER_{num}$ & $ER_{punc}$ & $\mathrm{WER_S}$ & WER & $ER_{lex}$ & $ER_{ent}$ & $ER_{num}$ & $ER_{punc}$ & $\mathrm{WER_S}$ & WER \\
\midrule
Hindi & IndicWhisper & 23.80 & 1.06 & 6.87 & 31.73 & 35.20 & 45.42 & 3.83 & 2.23 & 8.70 & 60.18 & 66.37 \\
& IndicConformer & \textbf{10.16} & 1.35 & 6.99 & 18.50 & 21.70 & \textbf{10.59} & 0.67 & 2.56 & 8.70 & 22.52 & 26.32 \\
& \textit{SCRIBE-ASR} & 11.68 & \textbf{0.31} & \textbf{3.30} & \textbf{15.29} & \textbf{17.57} & 8.58 & \textbf{0.59} & \textbf{0.59} & \textbf{6.73} & \textbf{16.49} & \textbf{19.29} \\
\midrule
Kannada & IndicWhisper & 19.29 & 2.06 & 10.09 & 31.44 & 40.51 & 17.99 & 1.16 & 3.03 & 12.46 & 34.64 & 46.09 \\
& IndicConformer & \textbf{12.46} & 2.49 & 10.29 & 25.24 & 32.95 & \textbf{15.13} & \textbf{0.87} & 3.96 & 12.46 & 32.42 & 40.74 \\
& \textit{SCRIBE-ASR} & 16.27 & \textbf{0.56} & \textbf{5.79} & \textbf{22.62} & \textbf{29.87} & 16.12 & 1.86 & \textbf{0.15} & \textbf{9.02} & \textbf{27.15} & \textbf{38.20} \\
\midrule
Malayalam & IndicWhisper & 14.65 & 1.74 & 15.41 & 31.80 & 41.77 & 17.76 & 1.52 & 1.58 & 14.29 & 35.15 & 54.74 \\
& IndicConformer & \textbf{13.58} & 2.39 & 15.40 & 31.37 & 41.00 & 17.32 & 1.39 & 3.67 & 14.29 & 36.67 & 52.11 \\
& \textit{SCRIBE-ASR} & 14.77 & \textbf{0.59} & \textbf{14.03} & \textbf{29.39} & \textbf{36.65} & \textbf{15.96} & \textbf{1.28} & \textbf{0.94} & \textbf{12.12} & \textbf{30.30} & \textbf{44.52} \\
\bottomrule
\end{tabular}
\vspace{-0.4cm}
\end{table*}

\section{Results}
\label{sec:results}

\subsection{Correlation with Human Judgment}

\begin{table}[h!]
\centering
\caption{Spearman $\rho$ correlation of SCRIBE error rates vs.\ monolithic WER with human expert ratings. SCRIBE's category-specific rates show consistent alignment ($|\rho| = 0.36$--$0.92$); global WER fails to significantly correlate with human judgment in several dimensions, particularly in Malayalam.}
\label{tab:correlation}
\footnotesize
\begin{tabular}{@{}llccc@{}}
\toprule
\textbf{Metric} & \textbf{Human Rating} & \textbf{Hi} & \textbf{Kn} & \textbf{Ml} \\ \midrule
$ER_{lex}$      & Lexical      & $-0.55$ & $-0.48$ & $-0.36$ \\
$ER_{num}$      & Numeral      & $-0.63$ & $-0.83$ & $-0.92$ \\
$ER_{punc}$     & Punctuation  & $-0.68$ & $-0.64$ & $-0.64$ \\ \cmidrule(l){1-5}
WER             & Lexical      & $-0.35$ & $-0.49$ & $-0.18$\textsuperscript{\dag} \\
WER             & Numeral      & $-0.61$ & $-0.40$ & $-0.03$\textsuperscript{\dag} \\
WER             & Punctuation  & $-0.32$ & $-0.49$ & $-0.16$\textsuperscript{\dag} \\ \bottomrule
\multicolumn{5}{l}{\textsuperscript{\dag}\scriptsize $p > 0.05$ indicating no statistically significant correlation.}
\end{tabular}
\vspace{-0.6cm}

\end{table}

Table~\ref{tab:correlation} confirms that SCRIBE's categorical metrics align robustly with human judgment ($|\rho| \!=\! 0.36$--$0.92$), significantly outperforming monolithic WER ($|\rho| \!\leq\! 0.49$). The alignment is strongest in high-stakes numeral accuracy, reaching $\rho \!=\! -0.92$ in Malayalam. Crucially, while WER fails to achieve statistical significance in several Malayalam dimensions ($p \!>\! 0.05$), SCRIBE's components remain highly predictive ($p \!\leq\! 0.001$). This disparity proves that experts prioritize functional categories—specifically punctuation and numerals—that global WER treats as noise.

Variations in lexical correlation ($|\rho| \!=\! 0.36$ in Malayalam to $0.55$ in Hindi) reflect the linguistic complexity of the evaluation set. The moderate alignment in Malayalam likely stems from its agglutinative nature, which increases subjectivity in human-perceived word boundaries. Nevertheless,  consistent significance of SCRIBE across all categories and languages underscores the need for multi-dimensional evaluation in specialized domains.

\vspace{-0.2cm}
\subsection{Diagnostic Decomposition}

Table~\ref{tab:combined_results} provides the categorical decomposition across general and out-of-distribution (OOD) legal benchmarks. While SCRIBE-ASR yields the lowest WER in all conditions, the diagnostic vector $\mathbf{E}$ reveals that the composition of these gains differs fundamentally across error categories.

\textbf{The WER inflation gap.} The most striking diagnostic finding appears in the Malayalam Legal set. WER reports 44.52\%, yet SCRIBE's decomposition reveals that genuine lexical failure ($ER_{lex}$) accounts for only 15.96\%.  By resolving valid morphological merges, SCRIBE's alignment engine reduces reported error inflation by roughly 30\% relative in both Malayalam and Kannada. An example showing 100\% WER on JIWER with 0\% Lexical error in SCRIBE is shown in Figure \ref{fig:alignment_comparison}.

Without this decomposition, the model would be deemed unusable based on WER alone; SCRIBE reveals that core acoustic-phonetic reliability is nearly 3$\times$ higher than the monolithic scalar suggests, and that a substantial portion of reported Indic WER is an artifact of morphological structure rather than acoustic misrecognition.

\textbf{Formatting generalization.} The most prominent model-level result is near-saturation of numeral formatting ($ER_{num} < 1\%$), with 75--96\% relative reduction compared to the best baseline across all benchmarks. Domain entity error ($ER_{ent}$) remains below 2\% even in OOD legal dictation, indicating that acoustic learning from general corpora transfers to specialized vocabulary. This generalization highlights the effectiveness of the LLM curation pipeline in producing training data whose formatting conventions extend to unseen domains.

\textbf{Punctuation as the remaining bottleneck.} Despite gains across formatting categories, $ER_{punc}$ remains the primary challenge, particularly in Dravidian languages. Malayalam Legal reports 12.12\% vs.\ Hindi's 6.73\%, a disparity visible only through categorical analysis. In agglutinative contexts, the prevalence of long compound wordforms makes prosodic segmentation harder to learn than numeral or entity formatting, pointing to prosodic modeling as the next development target.

\vspace{-0.2cm}
\subsection{SCRIBE as a Development Signal}

SCRIBE's diagnostic value extends beyond post-hoc evaluation to active model development, as illustrated by the feedback loop in Figure~\ref{fig:pipeline}. During training, early iterations exhibited systematic over-punctuation bias entirely invisible in aggregate WER, which improved monotonically. SCRIBE's $ER_{punc}$ decomposition isolated the regression to segments where legacy verbatim corpora contained extremely short sequences ($<$4 words) with misleading terminal punctuation, enabling targeted filtering and refined LLM curation prompts.

\vspace{-0.2cm}
\section{Conclusion}
\label{sec:conclusion}

Standard WER is insufficient on its own for rich transcription ASR: it provides no diagnostic signal and structurally penalizes agglutinative languages through cascading alignment failures. We introduced SCRIBE to address both through \textit{sandhi}-tolerant alignment and categorical error decomposition that complements rather than discards the global WER signal, validated by strong agreement with expert linguists. Our diagnostic analysis reveals a critical divergence: while formatting logic for numerals and entities generalizes effectively across domains, punctuation placement in agglutinative contexts remains the primary bottleneck. By resolving \textit{sandhi}-induced error inflation, SCRIBE proves that Indic ASR systems are more acoustically reliable than standard scalars suggest. We release SCRIBE alongside our curation recipe, benchmarks, and open-weight models to enable development of ASR systems meeting the correction thresholds required for professional dictation.

\section{Generative AI Use Disclosure}

The authors utilized large language model (LLM) tools, specifically Gemini 2.5 Pro, to facilitate the automated curation of rich transcription datasets (Section 4.1) and to assist in the linguistic refinement and technical polishing of the manuscript. All final content was reviewed, verified, and approved by the authors, who take full responsibility for the integrity of the research and its presentation.

\section{Acknowledgements}

We gratefully acknowledge the creators of the open-source Indic speech datasets that made this work possible. We thank Kriti Agarwal, Data operations lead, for coordinating the annotation effort, and the expert linguists who prepared the IN22-Legal ground truth and rated SCRIBE-ASR hypotheses: Jibin, Aswathi, Anusha, Rohith, Sanchayeeta, Abhilasha and Saumyaranjan.

\bibliographystyle{IEEEtran}
\bibliography{mybib}

\end{document}